\documentclass[sigconf]{acmart}

\AtBeginDocument{%
  }

\setcopyright{acmlicensed}

\copyrightyear{2024} 
\acmYear{2024} 
\setcopyright{acmlicensed}\acmConference[MM '24]{Proceedings of the 32nd
ACM International Conference on Multimedia}{October 28-November 1,
2024}{Melbourne, VIC, Australia}
\acmBooktitle{Proceedings of the 32nd ACM International Conference on
Multimedia (MM '24), October 28-November 1, 2024, Melbourne, VIC, Australia}
\acmDOI{10.1145/3664647.3681192}
\acmISBN{979-8-4007-0686-8/24/10}

\usepackage{utfsym}
\usepackage{multirow}
\usepackage{balance}
\usepackage{colortbl} % 加载colortbl包
\definecolor{mygray}{gray}{0.9}
\begin{document}

%%
%% The "title" command has an optional parameter,
%% allowing the author to define a "short title" to be used in page headers.
\title{Language-Driven Interactive Shadow Detection}

%%
%% The "author" command and its associated commands are used to define
%% the authors and their affiliations.
%% Of note is the shared affiliation of the first two authors, and the
%% "authornote" and "authornotemark" commands
%% used to denote shared contribution to the research.
\author{Hongqiu Wang}
\affiliation{%
    \institution{The Hong Kong University of Science and Technology (Guangzhou)}
    \country{Guangzhou, China}
}
\orcid{0000-0001-9726-4253}
\email{hwang007@connect.hkust-gz.edu.cn}

\author{Wei Wang}
%\authornote{Both authors contributed equally to this research.}
\affiliation{%
    \institution{University of Electronic Science and Technology of China}
    \country{Chengdu, China}
}
\orcid{0009-0003-5272-5062}
\email{202252012137@std.uestc.edu.cn}

\author{Haipeng Zhou}
%\authornote{Both authors contributed equally to this research.}
\affiliation{%
    \institution{The Hong Kong University of Science and Technology (Guangzhou)}
    \country{Guangzhou, China}
}
\orcid{0000-0001-5398-7847}
\email{hzhou321@connect.hkust-gz.edu.cn}

\author{Huihui Xu}
%\authornote{Both authors contributed equally to this research.}
\affiliation{%
    \institution{The Hong Kong University of Science and Technology (Guangzhou)}
    \country{Guangzhou, China}
}
\orcid{0009-0006-7868-1100}
\email{hxu047@connect.hkust-gz.edu.cn}

\author{Shaozhi Wu}
%\authornote{Both authors contributed equally to this research.}
\affiliation{%
    \institution{University of Electronic Science and Technology of China}
    \country{Chengdu, China}
}
\orcid{0000-0001-5466-8119}
\email{wszfrank@uestc.edu.cn}

% \author{Lei Zhu}
% \authornote{Corresponding author.}
% \affiliation{%
%     \institution{The Hong Kong University of Science and Technology (Guangzhou)}
%     \country{Guangzhou, China}
% }
% \email{leizhu@ust.hk}
\author{Lei Zhu}
\affiliation{%
  \institution{The Hong Kong University of Science and Technology (Guangzhou) \& The Hong Kong University of Science and Technology}
    \country{China}
 }
\email{leizhu@ust.hk}
\orcid{0000-0003-3871-663X}
\authornote{Corresponding author.}

%%
%% By default, the full list of authors will be used in the page
%% headers. Often, this list is too long, and will overlap
%% other information printed in the page headers. This command allows
%% the author to define a more concise list
%% of authors' names for this purpose.
\renewcommand{\shortauthors}{Hongqiu Wang et al.}

%%
%% The abstract is a short summary of the work to be presented in the
%% article.
\begin{abstract}
Traditional shadow detectors often identify all shadow regions of static images or video sequences. This work presents the Referring Video Shadow Detection (RVSD), which is an innovative task that rejuvenates the classic paradigm by facilitating the segmentation of particular shadows in videos based on descriptive natural language prompts. This novel RVSD not only achieves segmentation of arbitrary shadow areas of interest based on descriptions (\textbf{\textit{flexibility}}) but also allows users to interact with visual content more directly and naturally by using natural language prompts (\textbf{\textit{interactivity}}), paving the way for abundant applications ranging from advanced video editing to virtual reality experiences. To pioneer the RVSD research, we curated a well-annotated RVSD dataset, which encompasses 86 videos and a rich set of 15,011 paired textual descriptions with corresponding shadows. To the best of our knowledge, this dataset is the first one for addressing RVSD. Based on this dataset, we propose a Referring Shadow-Track Memory Network (RSM-Net) for addressing the RVSD task. In our RSM-Net, we devise a Twin-Track Synergistic Memory (TSM) to store intra-clip memory features and hierarchical inter-clip memory features, and then pass these memory features into a memory read module to refine features of the current video frame for referring shadow detection. We also develop a Mixed-Prior Shadow Attention (MSA) to utilize physical priors to obtain a coarse shadow map for learning more visual features by weighting it with the input video frame. Experimental results show that our RSM-Net achieves state-of-the-art performance for RVSD with a notable Overall IOU increase of 4.4\%. Our code and dataset are available at \textit{https://github.com/whq-xxh/RVSD}.
\end{abstract}

%%
%% The code below is generated by the tool at http://dl.acm.org/ccs.cfm.
%% Please copy and paste the code instead of the example below.
%%


%%
%% Keywords. The author(s) should pick words that accurately describe
%% the work being presented. Separate the keywords with commas.
%%
%% This command processes the author and affiliation and title
%% information and builds the first part of the formatted document.
\maketitle

\section{Introduction}
\label{sec:intro}
%\vspace{-3mm}
\begin{figure}[h]
	\centering
	\includegraphics[width=0.9\linewidth]{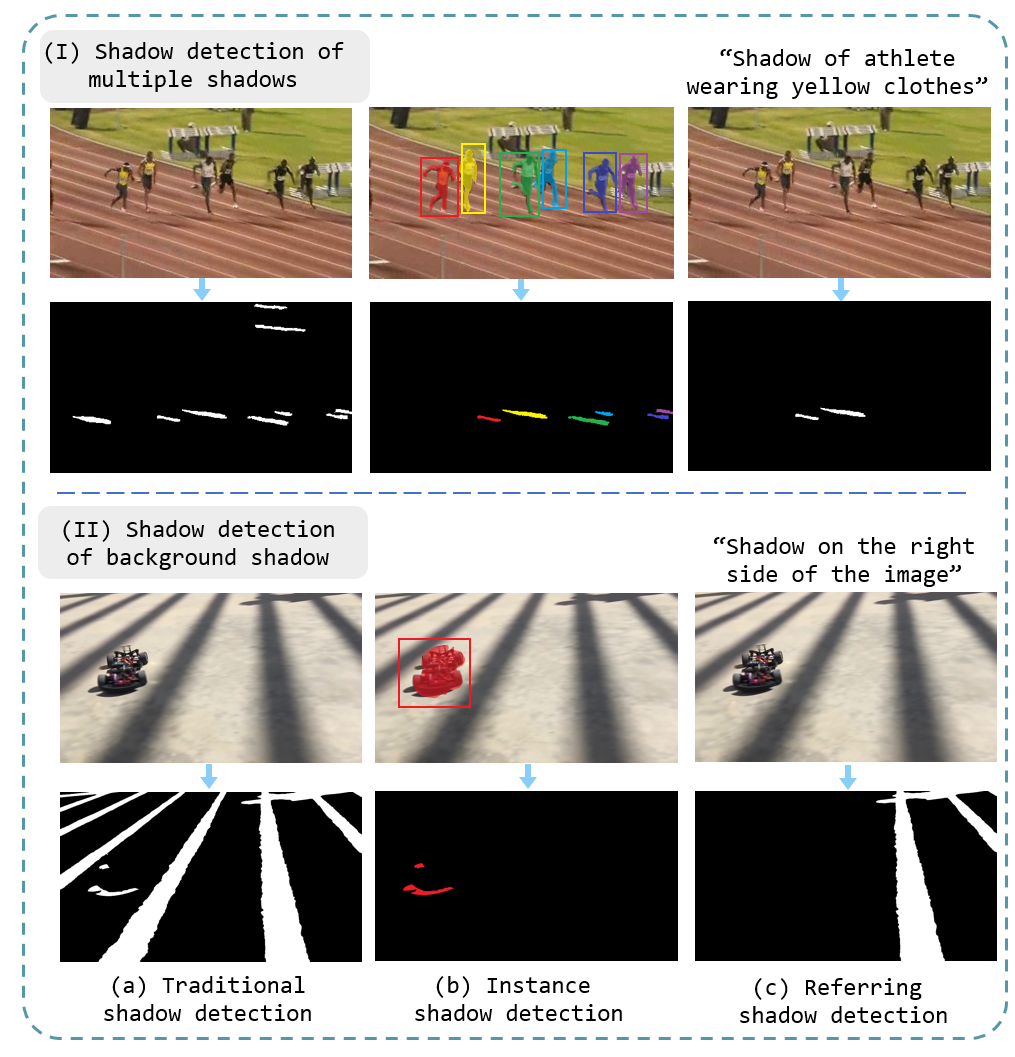}
	\vspace{-5mm}
	\caption{Comparisons of task settings for video shadow detection, instance shadow detection, and our RVSD. Traditional shadow detection (a) segments all shadows, current instance shadow detection (b) detects the foreground objects and segments the associated shadows, while our RVSD (c) can flexibly segment any shadow of interest referred by the text description, including I. those of multiple objects and II. background shadows (objects are invisible in the figure).}
	\label{fig:f1}
	\vspace{-3mm}
\end{figure}

As a cornerstone in computer vision, shadow detection has seen decades of research, and plays a pivotal role in interpreting the geometry and depth of a scene.
Accurate shadow detection can significantly enhance downstream computer vision tasks, including object detection, tracking, and scene reconstruction \cite{wehrwein2015shadow,he2021unsupervised,lyu2021sogan}. Previous shadow detection methods mainly focus on discerning all shadows in static images or video sequences and have achieved promising results \cite{zhu2018bidirectional,liao2021shadow,8578876,fang2021robust,zhu2022single,hou2021multiview,cong2023sddnet,zhou2024timeline}. 
However, these methods do not allow for the segmentation of specific shadows described by users, due to their lack of interactivity and flexibility which is essential in the era of advanced multimedia interaction.

Therefore, we propose Referring Video Shadow Detection (RVSD) as a novel task dedicated to meeting this demand by interactively segmenting the specific shadows in videos based on descriptive natural language prompts, as illustrated in Fig.~\ref{fig:f1}. 
The significance of RVSD can be envisaged in numerous applications. Advanced video editing tools can benefit from this RVSD task by allowing editors to precisely manipulate shadows based on verbal instructions \cite{chuang2003shadow,le2020shadow}. 
In augmented and virtual reality environments, understanding and reacting to user-specified shadows can enhance immersion and realism \cite{liu2020real,alhakamy2020real}. 
Furthermore, remote sensing may also benefit from this RVSD task to achieve more precise results \cite{liu2023rotated}.

\begin{figure*}[t]
	\centering
	\includegraphics[width=0.86\linewidth]{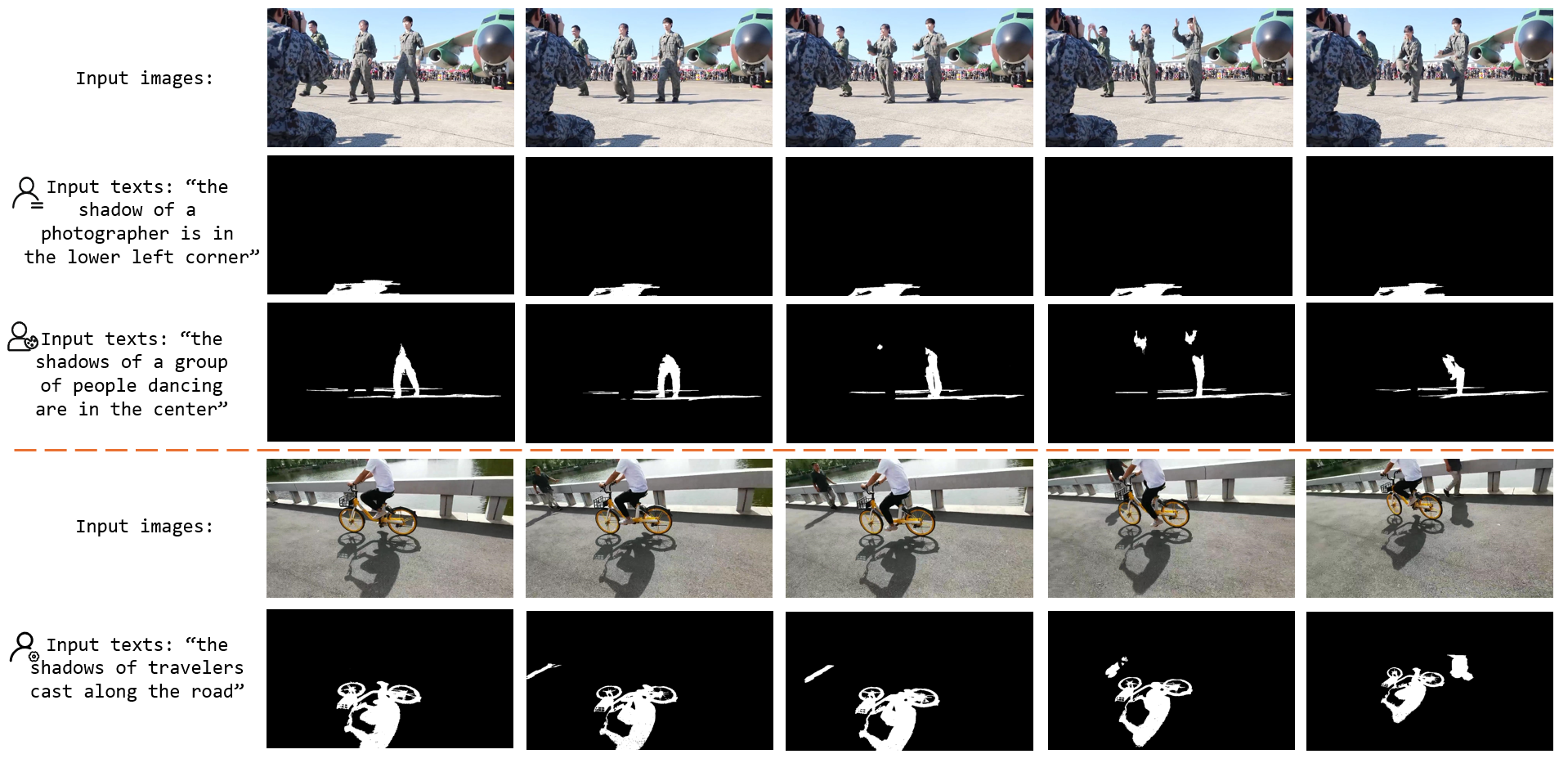}
	\vspace{-4mm}
	\caption{Sample frames from the RVSD dataset showcase pixel-level shadow annotations paired with textual descriptions that guide the corresponding shadow segmentation. These examples demonstrate that our RVSD not only facilitates the flexible segmentation of a specific shadow but also effectively segments shadows cast by groups of objects.}
	\label{fig:f2}
	\vspace{-3mm}
\end{figure*}

In this work, we introduce the first dataset dedicated to the RVSD task. This dataset collected 86 videos with 15,011 paired rich textual descriptions and corresponding shadow mask annotations, with some examples showcased in Fig.~\ref{fig:f2}. 
It has been meticulously curated to cover a wide array of scenarios, and shadow dynamics, which can provide a comprehensive benchmark for evaluating future RVSD methods. To the best of our knowledge, our dataset is the first one for RVSD with high-quality annotation.
%However, reaching an accurate RVSD is challenging. 

Compared to classical referring video object segmentation  \cite{botach2022end,seo2020urvos}, achieving an accurate RVSD is more challenging. Firstly, shadows lack rich appearance features and are easily confused with the dark regions of the input video~\cite{chen2021triple}. 
Moreover, shadows are naturally influenced by ever-changing environmental factors, demonstrating severe temporal shape transformations. 
Hence, RVSD requires richer temporal information \cite{wudesnow,wu2024rainmamba} and physical prior information to accurately identify the specific shadow referred by users' descriptions.

Based on our annotated RVSD dataset, we develop a novel Referring Shadow-Track Memory Network (RSM-Net) for addressing the RVSD task.
In our RSM-Net, we devise a Twin-Track Synergistic Memory (TSM) to store hierarchical inter-clip features and intra-clip features, and then propagate these two memory features via a memory read module to refine the current video frame for video shadow detection.
%improve the model's understanding of temporal dynamics, maintain referring shadow state consistency, and mitigate errors and drift in segmentation results. 
%%
Moreover, we propose a Mixed-Prior Shadow Attention (MSA) module that leverages physical knowledge to generate a preliminary shadow map, guiding the network to focus on potential shadow regions attentively.

Our main contributions are summarized as follows:
\vspace{-2mm}
\begin{itemize}
	\item This work is the first one to explore the RVSD task, which presents \textit{\textbf{a fresh paradigm for language-driven interactive and flexible shadow detection}} with potential benefits for numerous downstream tasks.
	\item To address the RVSD task, we collect and annotate the first dataset for RVSD consisting of 86 videos with 15,011 paired video frames and the corresponding text descriptions. This is the first dataset dedicated for the RVSD task.
	%We have produced and made publicly available the first RVSD dataset, aiming to foster further research within the related community.
	\item We propose an RSM-Net for RVSD. Here, we devise a TSM module to learn intra-clip and inter-clip features and store both of them in a memory to refine the current video frame for video shadow detection.
	Moreover, an MSA module is developed to generate the coarse shadow map for focusing on potential shadow areas for RVSD.
	\item Extensive experimental results show that our RSM-Net clearly outperforms state-of-the-art methods for the RVSD task.
\end{itemize}

\section{Related work}
\label{sec:related}

%-------------------------------------------------------------------------
% \vspace{3pt}\noindent\textbf{Shadow detection.} 
\subsection{Shadow detection.} 
\vspace{-1mm}
Shadow detection is crucial in computer vision, aiming to generate binary masks for all shadows \cite{liao2021shadow,8578876,fang2021robust,zhu2022single,hou2021multiview,cong2023sddnet,wang2021single}. Early techniques employed physical illumination and color models to analyze spectral and geometrical shadow properties \cite{panagopoulos2011illumination,tian2016new,salvador2004cast}. As the machine learning development, subsequent approaches built models using handcrafted attributes like texture \cite{zhu2010learning,vicente2015leave}, color \cite{lalonde2010detecting,guo2011single}, and edges \cite{lalonde2010detecting,huang2011characterizes}. These models were then combined with classifiers like decision trees \cite{lalonde2010detecting,zhu2010learning} and support vector machines \cite{guo2011single,huang2011characterizes,vicente2015leave} to differentiate shadow. However, the limited representational ability often restricted their effectiveness in various scenarios. 

Recent deep-learning methods have made significant strides in shadow detection by effectively learning from shadow images. Khan \textit{et al.} \cite{khan2014automatic} first propose a framework where relevant features are automatically learned through multiple convolutional deep neural networks. Shen \textit{et al.} \cite{7298818} employ structured CNNs to analyze shadow edges' local features. Vicente \textit{et al.} \cite{vicente2016large} recommend utilizing quickly obtained, partially accurate image labels, refining them automatically for enhanced performance. Hu \textit{et al.} \cite{8578876} develop a network using spatial RNNs for direction-aware context analysis in shadow detection. Chen \textit{et al.} \cite{chen2020multi} propose a semi-supervised, multi-task model combining consistency loss from unlabeled data with supervised loss from labeled data.

All the above investigations primarily focus on shadow detection for images, while there have been recent endeavors in video-based shadow detection as well. Chen \textit{et al.} \cite{chen2021triple} curate a new video shadow detection dataset and develop a triple-cooperative network for enhanced accuracy. Liu \textit{et al.} \cite{liu2023scotch} introduce shadow deformation attention trajectory, a new video self-attention module meticulously crafted to tackle substantial shadow deformations within videos. 
Differing from conventional video shadow detection that generates a universal binary shadow mask, we delve into a pioneering realm of referring video shadow detection. This approach facilitates the segmentation of each specific shadow through associated language expressions, enhancing user-friendly and personalized applications.

\vspace{-2mm}
\subsection{Referring segmentation.} 
\vspace{-1mm}
The aim of referring segmentation is to precisely outline the particular object referred to in a natural language expression within an image. This task combines computer vision and natural language processing, similar to our RVSD approach. Hu \textit{et al.} \cite{hu2016segmentation} first tackle the novel challenge of image segmentation guided by natural language expressions. They utilize a hybrid recurrent-convolutional network, encoding expressions with recurrent neural networks and generating response maps using fully convolutional networks from images. Liu \textit{et al.} \cite{liu2017recurrent} present a convolutional multimodal LSTM to encode interactions among words, visual, and spatial information components, enforcing a more effective word-to-image interaction.

\vspace{-2mm}
\subsection{Instance shadow detection.}
\vspace{-1mm}
Instance shadow detection is a subtask of shadow detection, and the goal of the current instance shadow detection is to detect the object and segment the corresponding shadow region.
Wang \textit{et al.} \cite{wang2020instance} pioneer instance shadow detection and built an image-based dataset for finer segmentation. Xing \textit{et al.} \cite{xing2022video} expand the scope of instance shadow detection from static images to dynamic videos, introducing a new framework to extract shadow-object associations in videos with paired tracking.

However, our RVSD stands out from previous approaches in two distinctions. Firstly, RVSD is a novel \textbf{\textit{interactive}} technique for shadow segmentation, which is more user-friendly. RVSD receives the user's linguistic description and the corresponding video frames as input to segment the relevant shadows, whereas instance shadow detection only inputs image data. Secondly, in contrast with previous approaches that focus on detecting objects in the foreground and segmenting their associated shadows separately, RVSD offers greater \textbf{\textit{flexibility}} by enabling the segmentation of any shadow of interest, including those of multiple objects or background shadows (objects are invisible), as illustrated in Fig.~\ref{fig:f1}.

\section{RVSD dataset} 
In this section, we introduce the newly established RVSD dataset. We begin by detailing the video collection, annotation and validation process in Section~\ref{3.1}, followed by an examination of dataset statistics and analysis in Section~\ref{3.2}.

\begin{table*}[t]
	\caption{Scene analysis of video shadows and corresponding linguistic description approaches. ✐ represents what we recommend as a description, $\circ$ represents the optional description, and $-$ represents no description.}
	\label{tab1}
	\vspace{-4mm}
	\begin{center}
		\resizebox{0.9\textwidth}{!}{
			\begin{tabular}{c|c|c|c|c|c|c|c|c|c}
				\hline
				\multicolumn{4}{c|}{Tag combination} & shadow & object & motion & shape & position & shadow type\\
				\hline
				\multirow{8}{*}{Only shadow} & \multirow{2}{*}{One shadow} & Stable & soft or hard & $\usym{2710}$ & $-$ & $-$ & $\usym{2710}$ & $\circ$ & $\circ$\\ 
				\cline{3-10} 
				&  & Moving & soft or hard & $\usym{2710}$ & $-$ & $\usym{2710}$ & $\usym{2710}$ & $\circ$ & $\circ$\\
				\cline{2-10}
				& \multirow{6}{*}{Multiple shadows} & \multirow{2}{*}{Stable} & soft or hard & $\usym{2710}$ & $-$ & $-$ & $\circ$ & $\usym{2710}$ & $\circ$\\
				\cline{4-10}
				&  &   & soft and hard & $\usym{2710}$ & $-$ & $-$ & $\circ$ & $\usym{2710}$ & $\usym{2710}$\\
				\cline{3-10}
				&   & \multirow{2}{*}{Moving} & soft or hard & $\usym{2710}$ & $-$ & $\usym{2710}$ & $\circ$ & $\usym{2710}$ & $\circ$\\
				\cline{4-10}
				&  &   & soft and hard & $\usym{2710}$ & $-$ & $\usym{2710}$ & $\circ$ & $\usym{2710}$ & $\usym{2710}$\\
				\cline{3-10}
				&   & \multirow{2}{*}{Stable and Moving} & soft or hard &  $\usym{2710}$ & $-$ & $\usym{2710}$ & $\circ$ & $\usym{2710}$ & $\circ$\\
				\cline{4-10}
				&  &   & soft and hard &  $\usym{2710}$ & $-$ & $\usym{2710}$ & $\circ$ & $\usym{2710}$ & $\usym{2710}$\\
				\hline
				\multirow{8}{*}{Shadow and object} & \multirow{2}{*}{One shadow} &  Stable &  soft or hard & $\usym{2710}$ & $\usym{2710}$ &  $-$ & $\usym{2710}$ & $\circ$ & $\circ$\\ 
				\cline{3-10} 
				&  & Moving & soft or hard & $\usym{2710}$ & $\usym{2710}$ & $\usym{2710}$ & $\usym{2710}$ & $\circ$ & $\circ$\\
				\cline{2-10}
				& \multirow{6}{*}{Multiple shadows} & \multirow{2}{*}{Stable} & soft or hard & $\usym{2710}$ & $\usym{2710}$ &  $-$ & $\circ$ & $\usym{2710}$ & $\circ$\\
				\cline{4-10}
				&  &   & soft and hard & $\usym{2710}$ & $\usym{2710}$ &  $-$ & $\circ$ & $\usym{2710}$ & $\usym{2710}$\\
				\cline{3-10}
				&   & \multirow{2}{*}{Moving} & soft or hard & $\usym{2710}$ & $\usym{2710}$ & $\usym{2710}$ & $\circ$ & $\usym{2710}$ & $\circ$\\
				\cline{4-10}
				&  &   & soft and hard & $\usym{2710}$ & $\usym{2710}$ & $\usym{2710}$ & $\circ$ & $\usym{2710}$ & $\usym{2710}$ \\
				\cline{3-10}
				&  & \multirow{2}{*}{Stable and Moving} & soft or hard & $\usym{2710}$ & $\usym{2710}$ & $\usym{2710}$ & $\circ$ & $\usym{2710}$ & $\circ$\\
				\cline{4-10}
				&  &   & soft and hard & $\usym{2710}$ & $\usym{2710}$ & $\usym{2710}$ & $\circ$ & $\usym{2710}$ & $\usym{2710}$\\
				\hline
		\end{tabular}}
		\vspace{-3mm}
	\end{center}
\end{table*}

\subsection{Constructing the RVSD dataset}\label{3.1}
\subsubsection{Video Collection.} 
We perform data collection and re-annotation using the most extensive publicly available video shadow dataset to date, which comprises 120 videos, known as ViSha \cite{chen2021triple}. After reviewing 120 potential candidates, we carefully exclude shadows that are blended together (hard to differentiate them through textual descriptions). We also remove highly fragmented shadows. Meanwhile, we re-mask the original binary shadow mask to label the shadows of different instances separately. 
Eventually, with an emphasis on quality over quantity, we meticulously select 86 videos to construct a benchmark dataset that encompasses diversity and represents a broad spectrum of real-world video scenarios. 
We also conduct an analysis of various scenarios where shadows are present in the video. Further details can be found in Table~\ref{tab1}. According to the tag combinations for different shadow scenarios, we give reference suggestions for the corresponding language descriptions. The following section will provide a more detailed explanation of the language expression annotation and validation process.

\subsubsection{Language Description Annotation.} 
The fundamental methodology and procedure for language annotation in RVSD are following previous works~\cite{kazemzadeh2014referitgame,seo2020urvos,dong2022reading,ding2023mevis}. It employs an interactive approach involving multiple annotators taking turns to annotate and validate. We invited multiple people with no computer vision-related background to participate in labeling with the guidance in our Table~\ref{tab1}. The annotator is required to select one or multiple shadows from the video and generate corresponding referring descriptions according to the guidance for annotating language expressions. It is worth noting that our guidance in Table~\ref{tab1} contains three elements, the recommended $\usym{2710}$, optional $\circ$, and not required $-$ ones. In this setup, there is a distinct advantage. While ensuring the accurate depiction of various shadows, annotators have the freedom to select the description's content, thus maximizing the richness and flexibility of the sentences.

\subsubsection{Language Description Validation.} 
After the initial annotation, we conduct validation tasks for all annotations. The validation process commences by presenting the video along with the corresponding expression, prompting the validator to identify the shadow referred to in the expression. The validator is required to independently find the target shadow and record it. The shadow chosen by the validator is subsequently cross-referenced with the annotations provided by the annotator. If they align, the annotation passes the validation; otherwise, it undergoes re-labeling and subsequent validation. By implementing the validation procedure, we strive to ensure the accurate language representation of shadows within our dataset, while maintaining sentence diversity.

\begin{table}[t]
	\centering
	\caption{Detailed information of the proposed RVSD datasets.}
	\vspace{-3mm}
	\label{tab2}
	\resizebox{0.46\textwidth}{!}{%
		\begin{tabular}{c|c|c|c}
			\hline
			\multirow{2}{*}{Dataset} &\multicolumn{3}{c}{Paired prompt textual description} \\
			\cline{2-4} & Shadow-text pairs & Min word count & Max word count \\
			\hline
			{\centering RVSD} & {15,011 pairs} & {6 words} & {27 words} \\
			\hline
	\end{tabular}}
	\begin{flushleft}
		% \vspace{-1mm}
		\textbf{Example 1}: "the soft shadow is located below"
		
		\textbf{Example 2}: "the hard shadow of a person who is holding an umbrella and walking is in the upper left corner"
	\end{flushleft}
	\vspace{-5mm}
\end{table}

\subsection{Analysis and Statistics of the RVSD dataset}\label{3.2}

\subsubsection{Video and Shadow.} 
The selection of video and shadow is based on diversity, which makes the dataset both comprehensive and challenging. Firstly, the number of shadows is a factor to consider, since RVSD in a scene with multiple shadows generally is more challenging than one with a single shadow. Multiple shadows, often close in location and similar in shape, can lead to confusion during model interpretation, thereby making precise segmentation of a specific shadow challenging. In the RVSD dataset, there are 53 videos containing multiple shadows, and 33 videos containing only one shadow. Next, we consider the type of shadow, namely hard and soft shadow. A hard shadow has a distinct boundary, whereas a soft shadow's boundary might be fuzzy or indistinct. Soft shadow is more challenging to segment than hard shadow. The RVSD dataset includes 156 hard shadows and 42 soft shadows. Finally, the diversity of scenes is also an important factor, diverse scenes are closer to the real world, making the dataset more comprehensive. The RVSD dataset includes 74 daytime videos and 12 nighttime videos, as well as 24 indoor videos and 62 outdoor videos.

\begin{figure}[t]
	\centering
	\includegraphics[width=1\linewidth]{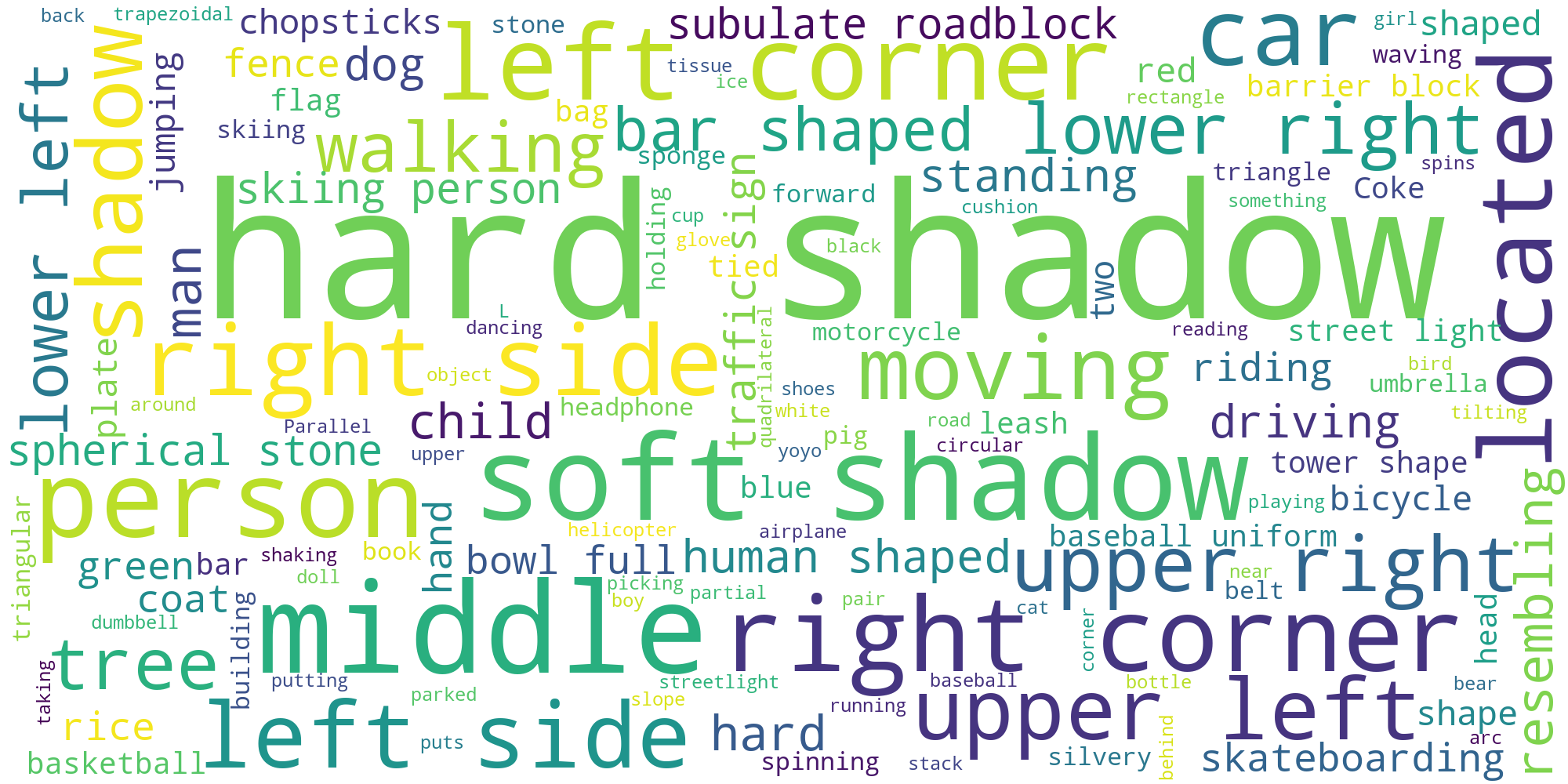}
	\vspace{-6mm}
	\caption{Word cloud of the RVSD dataset. The RVSD dataset encompasses a vast vocabulary that captures shadows from various perspectives, encompassing aspects like shadow type, location, shape, movement, and associated objects.}
	\label{fig:word}
	\vspace{-6mm}
\end{figure}

\begin{figure*}[t]
	\centering
	\includegraphics[width=1\linewidth]{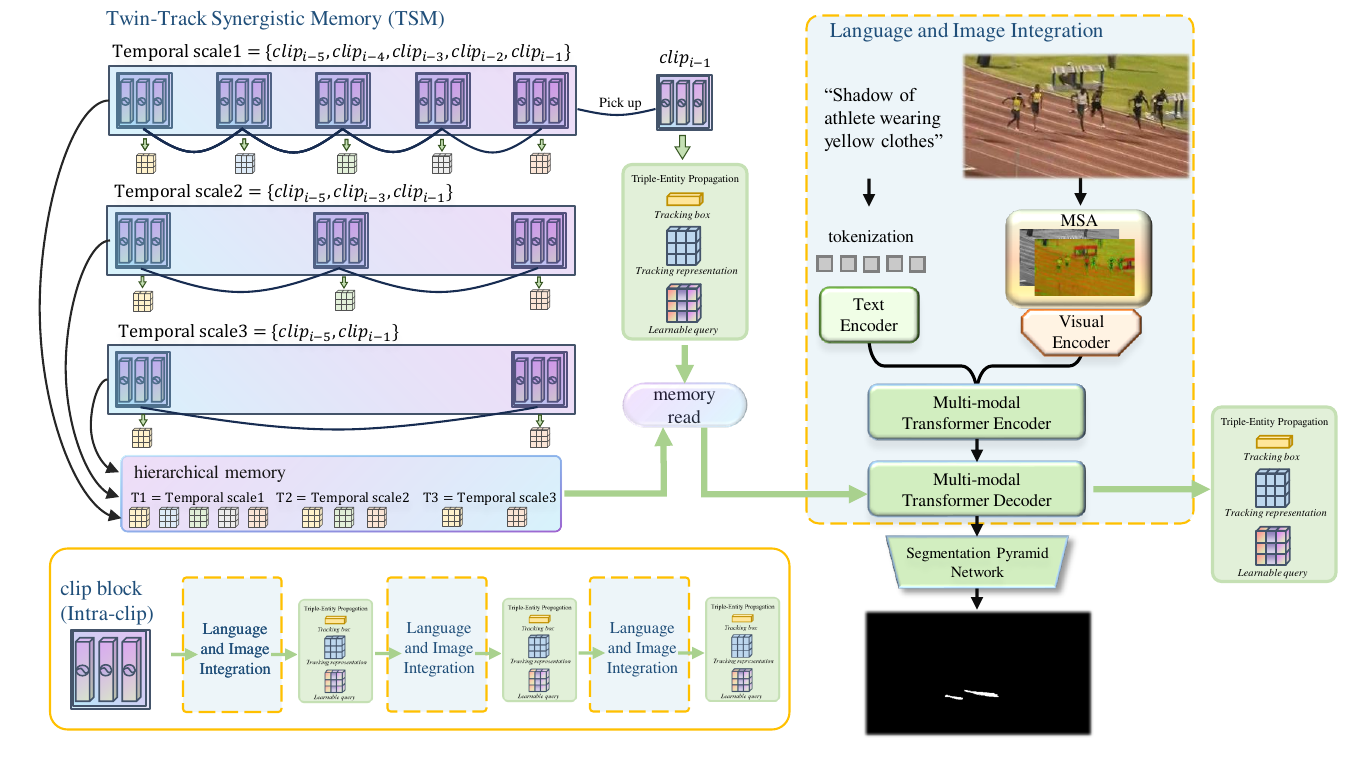}
	\vspace{-10mm}
	\caption{An overview of our approach. The TSM on the left side represents the construction phase of twin-track memory, which contains both inter-clip and intra-clip tracks of memory. The clip block (intra-clip) in the lower left corner signifies memory propagation between frames, with each clip in the figure containing three video frames (Language and Image Integration). Eventually, The hierarchical memory is strategically accessed for processing the current frame, ensuring comprehensive and context-aware shadow detection.}
	\label{fig:method}
	\vspace{-3mm}
\end{figure*}

\subsubsection{Language Description.} 
One significant characteristic of the RVSD dataset is assigning language descriptions to each shadow on a frame-by-frame basis. Firstly, we incorporate language descriptions for shadows based on the ViSha dataset. The preceding referring object segmentation dataset \cite{seo2020urvos} offered two kinds of language annotations: full-video expression and first-frame expression. The first-frame expression solely relies on the static attributes of the first frame image, whereas the full-video expression takes into account the whole video. However, neither of these expressions can perfectly align with the state of each target in the video (due to the movements of objects within the video), leading to instances where certain frames do not correspond with their language descriptions. Therefore, when adding language descriptions, we add corresponding language descriptions for each shadow target and its corresponding frames, making the content of the language descriptions match each frame. Meanwhile, we've established guidelines for the method of annotating language description, making the description more closely reflect the status of the target.
In the language descriptions, we include the type of shadow, followed by the static attributes, the position information of the shadow, and the action information of the object corresponding to the shadow. Table~\ref{tab2} presents the basic statistics of our language descriptions. Fig.~\ref{fig:word} displays the word cloud for RVSD, featuring terms associated with shadow type (e.g., “hard shadow”, “soft shadow”), actions of the shadow-casting subjects (e.g., “walking”, “skateboarding”), position (e.g., “left”, “upper right”), and the objects casting the shadows (e.g., "person" or "tree").

\section{Methodology}
\subsection{Overview}
Our approach is designed to generate a binary mask of the specified shadow of the given input video based on the corresponding natural language expression. Fig.~\ref{fig:method} presents a schematic overview of our RSM-Net, specifically devised for the RVSD task.
RSM-Net takes language descriptions and video frames as input, as depicted in the "Language and Image Integration" box to the right of Fig.~\ref{fig:method}. It reads the previously stored Twin-Track Synergistic Memory and processes the image through the MSA module to jointly achieve the accurate segmentation of the target shadow.

We first present how the twin-track synergistic memory is constructed and the content of our Triple-Entity memory propagation in Sec.~\ref{sec4.1}. Subsequently, Sec.~\ref{sec4.2} is dedicated to a comprehensive exposition of our language and image integration, encompassing aspects of hierarchical memory reading and memory propagation. Lastly, in Sec.~\ref{sec4.3}, we describe how our MSA effectively utilizes prior knowledge to facilitate the RVSD task.

\subsection{Twin-Track Synergistic Memory (TSM) Construction} \label{sec4.1}
In response to the dynamic nature of shadows, we incorporate a twin-track memory structure into our framework. This encompasses both past hierarchical inter-clip memory and intra-clip memory, established before processing the current frame. The integration of these sequential memories in our network allows for a nuanced capture of evolving information across the video sequence, significantly contributing to the precise segmentation of the current frame. We emphasize that our hierarchical memory is dynamically updated. Specifically, when processing the current frame $clip_i$, we constructed memory using the preceding five clips $\{clip_{i-5},...,clip_{i-1}\}$. This dynamic updating and utilization in memory enhances our network's flexibility and efficiency while avoiding the disturbances from significant variations in very early video frames, allowing it to continuously adapt to the varying characteristics of shadows during the segmentation process.

First, our hierarchical inter-clip memory is generated from multiple clip blocks. The details of the clip block are in the yellow box in the lower left corner of Fig.~\ref{fig:method}. It mainly contains triple-entity memory propagation. In the triple-entity, the $tracking\ box$ $t_{box}\in\mathbb{R}^{N_q\times 4}$ denotes the predicted shadow detection box, $tracking\ representation$  $t_{rep}\in\mathbb{R}^{N_q\times d}$ refers to the encoded multi-modal features, and $learnable\ query$ $t_{que}\in\mathbb{R}^{N_q\times d}$ signifies the shadow query. Our hierarchical inter-clip memory stores $t_{rep}s$ from three different time scales $\{\text{T1,T2,T3}\}$. Meanwhile, we pick up the triple-entity of the previous frame of the currently pending frame from $clip_{i-1}$ as the intra-clip memory. Next, we will elaborate on the processing of the current frame and the efficient retrieval of information from the twin-track memory.

\subsection{Referring Video Frame Shadow Segmentation and Memory Read} \label{sec4.2}
\begin{figure}[t] 
	\centering
	\includegraphics[width=1\linewidth]{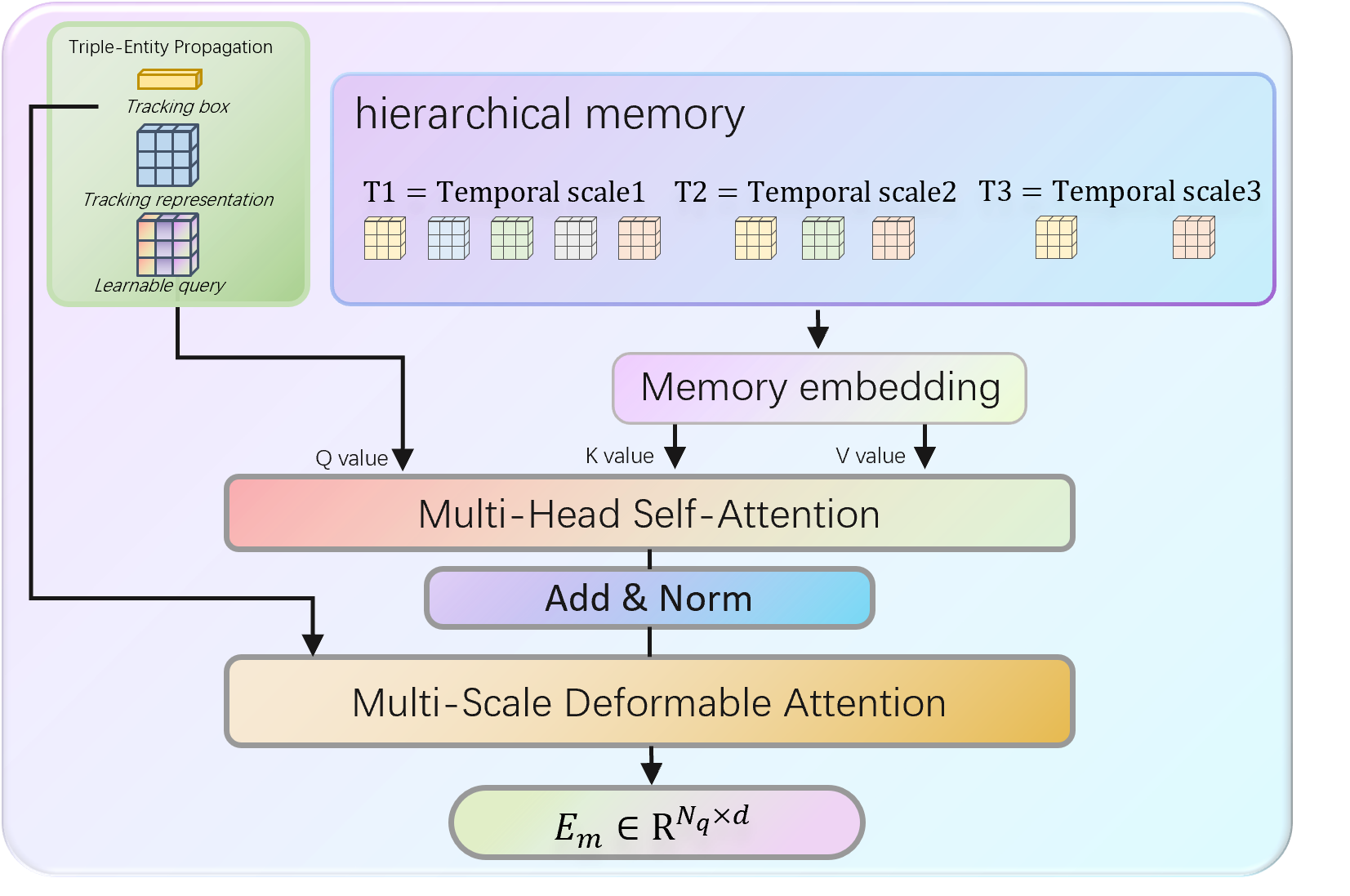}
	\vspace{-8mm}
	\caption{Details of hierarchical memory reading. First, perform memory embedding on it and then input it into the self-attention module together with the $\textbf{\textit{L}earnable\ query}$.}
	\label{fig:mem}
	\vspace{-4mm}
\end{figure}

In line with earlier researches \cite{zhu2020deformable,botach2022end,wu2022language}, our learnable query-based referring segmentation largely follows a well-established paradigm, the Deformable DETR \cite{zhu2020deformable}. This involves processing a video frame, a textual expression, and a set of learnable queries. The output includes the target bounding box, segmentation mask, and associated output embeddings that match the input language expression.

To process a particular $t_{th}$ frame image $I_t$, our network first performs attention enhancement of potential shadow regions via the MSA module, followed by extracting image feature $\mathcal{F}_i$ via the visual encoder. Meanwhile, the associated language description $\varepsilon=\{e_i\}_{i=1}^N$ with $N$ words, where $e_i$ is the $i_{th}$ word is tokenized and fed into the text encoder, yielding $\mathcal{F}_\varepsilon$.
Next, the visual and linguistic embeddings are linearly projected to a unified embedding space with the same dimension and concatenated to form a multi-modal embedding, denoted as $\mathcal{F}_m=\{\mathcal{F}_i,\mathcal{F}_\varepsilon\}$. This is then input into the transformer encoder, realizing cross-modal information fusion and interaction.

In the reading phase, it receives the triple-entity from the preceding frame along with the hierarchical memory, subsequently outputting the $E_m\in\mathbb{R}^{N_q\times d}$. Initially, we apply memory embedding to the hierarchical memory as shown in Fig.~\ref{fig:mem}, thereby fusing features across various temporal scales as follows:
\begin{equation}
	h_{\text{mem}} = \Phi_{\text{MLP}}^{(3)}\left( Concat\left( FC(\text{T1}), FC(\text{T2}), FC(\text{T3}) \right) \right),
\end{equation}
where $\Phi_{\text{MLP}}^{(3)}$ represents a three-layer MLP (Multi-Layer Perceptron). $Concat$ and $FC$ stand for concatenated operations and fully connected layers, respectively. The $h_{\text{mem}}$ is fed into the Multi-Head Self-Attention module as both the Key (K) and Value (V), while the Query (Q) originates from the intra-clip $learnable\ query$ $t_{que}$. During the propagation phase, the $t_{rep}$ advances the output as $\Phi_{\text{MLP}}^{(3)}(E_m)$.

To generate the final mask, our segmentation pyramid network implements a cross-modal FPN (Feature Pyramid Network) \cite{lin2017feature} to enable multi-scale fusion between the linguistic features and visual feature maps (see Sec.~\ref{sec5} for model details). To summarize, the training objective of our network is to minimize the loss function as follows:
\begin{equation}
	\underset{\alpha_{\text{box}}, \beta_{\text{mask}}}{\text{arg min}}\, \mathcal{L}_{\text{refer}} = \alpha_{\text{box}} \mathcal{L}_{\text{box}} + \beta_{\text{mask}} \mathcal{L}_{\text{mask}},
\end{equation}
where $\mathcal{L}_{\text{box}}$ denotes the loss associated with bounding boxes, which is a composite of L1 loss and GIoU loss \cite{rezatofighi2019generalized}. $\mathcal{L}_{\text{mask}}$ signifies the loss pertaining to masks, aggregating DICE loss \cite{milletari2016v} and the focal loss for binary masks. The coefficients for the losses are denoted as $\alpha_{\text{box}}$ for $\mathcal{L}_{\text{box}}$, and $\beta_{\text{mask}}$ for $\mathcal{L}_{\text{mask}}$.

\subsection{Mixed-Prior Shadow Attention (MSA)} \label{sec4.3}
\begin{figure}[t]
	\centering
	\includegraphics[width=1\linewidth]{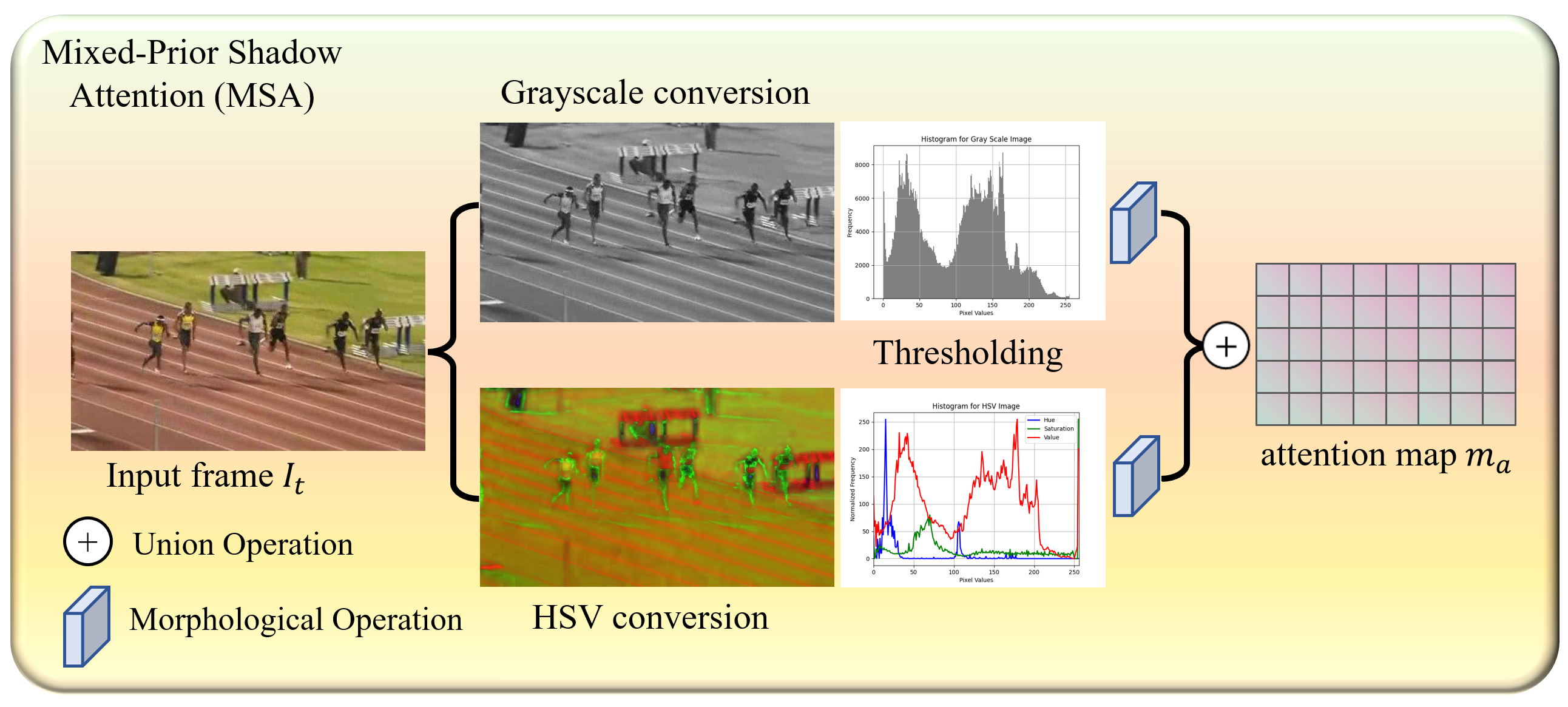}
	\vspace{-8mm}
	\caption{Illustration of MSA. The image is transformed into two distinct color spaces, enabling thresholding and morphological operations that result in the creation of a shadow attention map.}
	\label{fig:msa}
	\vspace{-4mm}
\end{figure}

The premise and foundation of the RVSD task is that the network can recognize and pay attention to the shadow region. Therefore, we design a Mixed-Prior Shadow Attention (MSA) module to utilize the physical prior knowledge \cite{finlayson2005removal,cucchiara2001improving} to generate a weight map $m_a$ to help the network focus more on the shadow region. Given the complexity and variability of the shadow background, our MSA identifies the shadow region from two different color spaces to improve robustness and generalization, as illustrated in Fig.~\ref{fig:msa}.

Grayscale Thresholding.
For a given image $I_t$, the grayscale representation is obtained by converting the RGB channels into a single intensity channel \( I_{\text{gray}} \). Mathematically, the binary shadow mask \( M_{\text{gray}} \) is obtained as:

\begin{equation}
	M_{\text{gray}} = \left\{ (x, y) | T_{\text{min}} \leq I_{\text{gray}}(x, y) \leq T_{\text{max}} \right\}, \label{e4}
\end{equation}
where the thresholds [$T_{\text{min}}$,~$T_{\text{max}}$] specify the acceptable range of grayscale values for identifying shadow regions.

\begin{table*}[t]
	\centering
	\caption{Quantitative comparison of our network and other state-of-the-art methods on the RVSD Dataset.}
	\vspace{-4mm}
	\label{tab3}
	\resizebox{0.9\textwidth}{!}{%
		\begin{tabular}{c|c| c c c c c|c c|c}
			\hline
			\rowcolor{mygray} &  & \multicolumn{5}{c|}{Precision} & \multicolumn{2}{c|}{IoU} & \\ 
			\rowcolor{mygray}
			\multirow{-2}{*}{Method}& \multirow{-2}{*}{Pub.} & P@0.5$\uparrow$ & P@0.6$\uparrow$ & P@0.7$\uparrow$ & P@0.8$\uparrow$ & P@0.9$\uparrow$ & Overall$\uparrow$ & Mean$\uparrow$ & \multirow{-2}{*}{mAP$\uparrow$}\\
			\hline
			\hline
			URVOS \cite{seo2020urvos} & ECCV 2020 & 51.2 & 44.8 & 37.5 & 28.6 & 16.1 & 58.5 & 45.9 & 33.5 \\
			CMPC-V \cite{liu2021cross} & TPAMI 2021 & 51.3 & 46.8 & 39.1 & 30.6 & 17.3 & 57.8 & 46.5 & 34.6 \\
			LBDT \cite{ding2022language} & CVPR 2022 & 55.0 & 50.3 & 40.6 & 31.1 & 17.6 & 62.6 & 48.9 & 36.4 \\
			MTTR \cite{botach2022end} & CVPR 2022 & 56.5 & 51.6 & 41.6 & 33.3 & 19.6 & 61.4 & 51.3 & 37.9 \\
			ReferFormer \cite{wu2022language} & CVPR 2022 & 70.1 & 64.2 & 56.5 & 45.1 & 26.8 & 67.2 & 61.2 & 49.3 \\
			SgMg \cite{Miao_2023_ICCV} & ICCV 2023 & 70.3 & 66.0 & 57.4 & 44.5 & 27.3 & 68.3 & 62.2 & 49.7 \\
			R2VOS \cite{li2023robust} & ICCV 2023 & 71.2 & 66.9 & 58.3 & 45.0 & 27.0 & 69.8 & 62.4 & 50.2 \\
			OnlineRefer \cite{wu2023onlinerefer} & ICCV 2023 & 71.5 & 65.7 & 57.8 & 46.1 & \textbf{27.9} & 70.2 & 62.5 & 50.5 \\
			\hline
			\hline
			Our RSM-Net & - & \textbf{73.1} & \textbf{68.0} & \textbf{61.2} & \textbf{46.8} & 27.1 & \textbf{74.6} & \textbf{64.3} & \textbf{51.8} \\
			Gain & - & $\uparrow$1.6 & $\uparrow$1.1 & $\uparrow$2.9 & $\uparrow$0.7 & $\uparrow$- & $\uparrow$4.4 & $\uparrow$1.8 & $\uparrow$1.3 \\
			\hline
	\end{tabular}}
	% \vspace{-3mm}
\end{table*}

\begin{table*}[t]
	\centering
	\caption{Ablation study of our methods on the RVSD Dataset. IntraC and InterC represent our network with only intra-clip memory features and only the inter-clip memory features, respectively. The InterC-S and InterC-H denote a single scale (Temporal scale 1) and a hierarchical scale (Temporal scale 1, 2, and 3), respectively for learning the inter-clip memory features.}\label{tab4}
	\vspace{-4mm}
	\resizebox{0.92\textwidth}{!}{%
		\begin{tabular}{c|c|c c c|c c c c c|c c|c}
			\hline
			\rowcolor{mygray}
			&  & \multicolumn{3}{c|}{TSM} & \multicolumn{5}{c|}{Precision} & \multicolumn{2}{c|}{IoU} & \\
			\rowcolor{mygray}
			\multirow{-2}{*}{Method} & \multirow{-2}{*}{MSA} & IntraC & InterC-S & InterC-H & P@0.5$\uparrow$ & P@0.6$\uparrow$ & P@0.7$\uparrow$ & P@0.8$\uparrow$ & P@0.9$\uparrow$ & Overall$\uparrow$ & Mean$\uparrow$ & \multirow{-2}{*}{mAP$\uparrow$}\\
			\hline
			\hline
			Baseline & - & - & - & - & 68.8 & 62.3 & 56.2 & 42.4 & 25.5 & 65.7 & 60.4 & 47.7 \\
			M1 & \checkmark & - & - & - &70.0 & 65.1 & 57.8 & 43.1 & 25.8 & 68.2 & 61.3 & 49.0 \\
			M2 & \checkmark  &\checkmark & - & - & 71.9 & 66.6 & 59.5 & 45.6 & 27.0 & 71.8 & 62.8 & 50.7  \\
			M3 & \checkmark  &\checkmark & \checkmark & - & 72.6 & 67.1 & 59.5 & 45.7 & \textbf{27.2} & 72.0 & 63.3 & 51.0 \\
			Our RSM-Net & \checkmark & \checkmark &\checkmark & \checkmark & \textbf{73.1} & \textbf{68.0} & \textbf{61.2} & \textbf{46.8} & 27.1 & \textbf{74.6} & \textbf{64.3} & \textbf{51.8} \\
			\hline
	\end{tabular}}
	\vspace{-3mm}
\end{table*}

HSV Thresholding.
The input frame $I_t$ is also converted to the HSV color space \( I_{\text{HSV}} \). In the HSV color space, shadow regions typically exhibit lower values in the Saturation ($S$) and Value ($V$) channels due to color desaturation and reduced brightness caused by occlusion. Thus, two thresholds \( T_{S} \) and \( T_{V} \) are applied on the Saturation and Value channels respectively to discern the shadow regions. The binary shadow mask \( M_{\text{HSV}} \) is obtained as:

\begin{equation}
	\begin{aligned}
		M_{HSV} = \left\{ (x, y) \right. & | T_{S_{\text{min}}} \leq S(x, y) \leq T_{S_{\text{max}}}, \\
		& \left. T_{V_{\text{min}}} \leq V(x, y) \leq T_{V_{\text{max}}} \right\}.    
	\end{aligned} \label{e5}
\end{equation}
The thresholds \( T_{S_{\text{min}}}, T_{S_{\text{max}}}, T_{V_{\text{min}}}, \) and \( T_{V_{\text{max}}} \) specify the acceptable range for shadow regions in the \( S \) and \( V \) channels.

The combined shadow mask is then obtained by:
\begin{equation}
	M_{\text{combined}}(x, y) = \Psi_{mor}(M_{\text{HSV}}(x, y)) \cup \Psi_{mor}(M_{\text{gray}}(x, y)).
\end{equation}
The morphological operation ($\Psi_{mor}$) here specifically refers to the "opening" operation function with a kernel of size \(5 \times 5\). This operation comprises an erosion followed by a dilation, which helps to remove noise and small interference while keeping the structure of the referring shadow. The final combined mask $M_{\text{combined}}$ is then obtained by taking the union (logical OR operation denoted by $\cup$) of these morphologically processed masks. The $M_{\text{combined}}$ is then weighted as attention map $m_a$ on the input frame $I_t$.

\section{Experiments} \label{sec5}

\subsection{Evaluation Metrics and Data Setting.} 
Following previous segmentation works\cite{botach2022end,ding2022language,wang2024video,wang2024dual}, we employ standard metrics for quantitative comparisons. These metrics are Precision@K, Overall IoU, Mean IoU, and mAP (mean Average Precision) across an IoU (Intersection over Union) range from 0.50 to 0.95 (a step of 0.05). 
IoU measures the overlap between predicted and ground truth regions, while precision@K evaluates the proportion of test instances surpassing the IoU threshold of K. 
The mAP metric computes the mean precision over varying IoU thresholds. 
%Generally, a superior RVSD method often has higher scores of these metrics.
To ensure a rich variety of scenarios in both the training and testing phases, the dataset is reasonably divided into training and testing subsets. The training subset encompasses 54 videos paired with 9,856 sentence-shadow combinations, while the testing subset contains 32 videos accompanied by 5,155 pairs.

\subsection{Implementation and Training Details.}
We utilize ResNet50 \cite{he2016deep,wang2023dynamic,wu2023mask} as the visual backbones for extracting features. For the text encoding, we employ RoBERTa \cite{liu2019roberta} and freeze its parameters \cite{yuan2024auformer} throughout the training process \cite{wang2023advancing,wang2024advancing}. Following \cite{zhu2020deformable,wu2023onlinerefer}, the final three-stage features from our visual backbone serve as inputs for both the Transformer \cite{lu2024gpt,liu2023multi} encoder and the FPN \cite{lin2017feature}. The encoder and decoder of our Multi-modal Transformer framework have four layers, operating at a dimensionality of $d = 256$.
The threshold range [\( T_{S_{\text{min}}}, T_{S_{\text{max}}}], [T_{V_{\text{min}}}, T_{V_{\text{max}}} \)] were empirically set as $[[0, 155], [6, 130]]$.

We perform all experiments using PyTorch on an NVIDIA GeForce RTX 3090 GPU with 24 GB of memory for training. Our model optimization employs the AdamW optimizer \cite{loshchilov2017decoupled} with an initial learning rate set at $1e-5$, while the visual backbone is adjusted at a lower rate of $5e-6$. The training spans 20 epochs, and the featuring learning rate reductions by a factor of 0.1 after the 3rd and 5th epochs. The initial frame's query number $N_q$ is set to 5. Each video frame is resized to ensure a minimum dimension of 320 on the shorter side and a maximum of 576 on the longer side.

\begin{figure*}[t]
	\centering
	\includegraphics[width=0.86\linewidth]{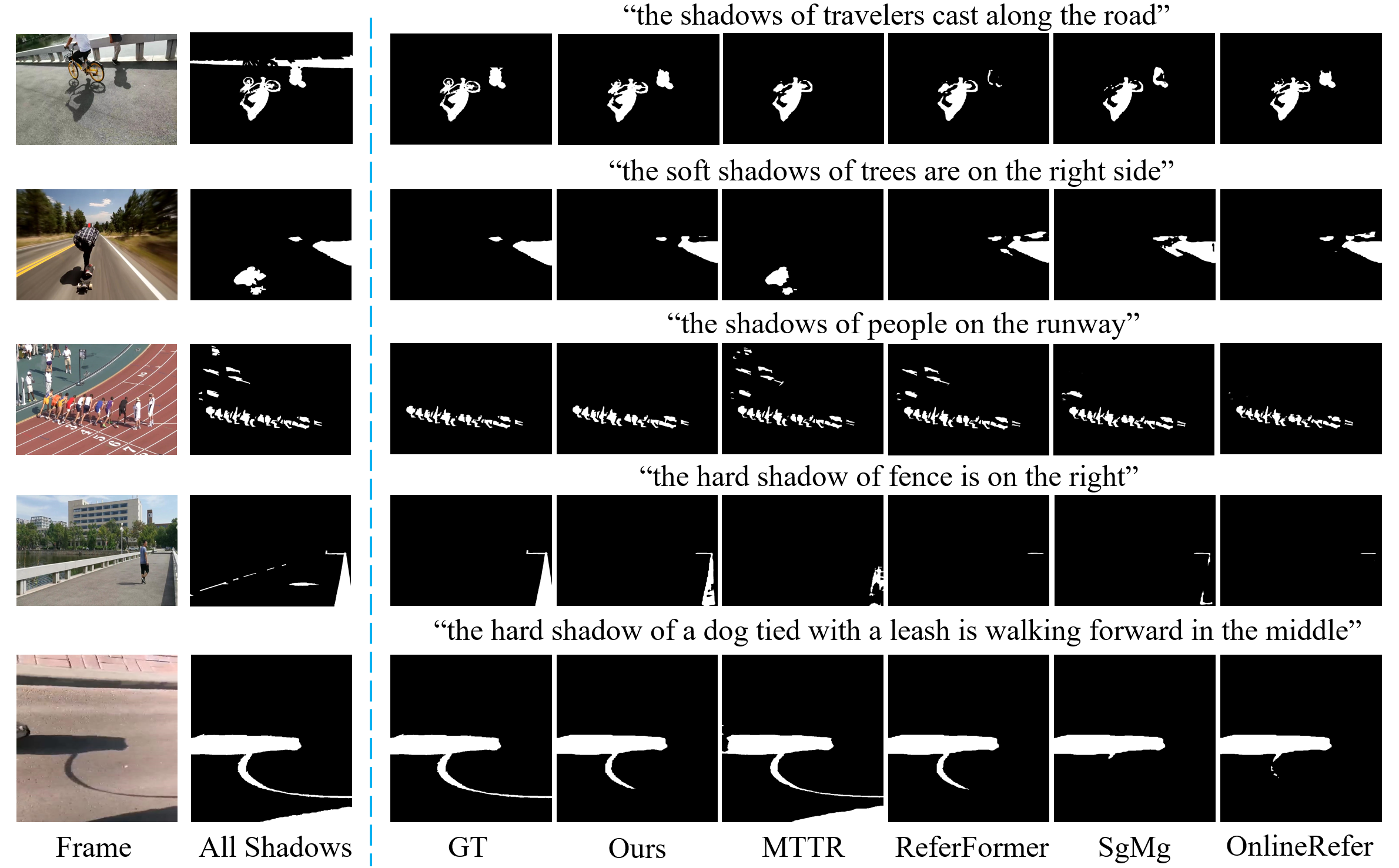}
	\vspace{-3mm}
	\caption{Visual comparisons of predicted video shadow detection results referring by the text descriptions. Apparently, our RSM-Net clearly outperforms compared methods and achieves more accurate shadow detection results based on the relevant text descriptions. Please zoom in for more details.}
	\label{fig:demo}
	\vspace{-2mm}
\end{figure*}

\subsection{Comparison with State-of-the-art Methods}
We compare our network against state-of-the-art methods on the RVSD dataset. 
These compared methods include URVOS \cite{seo2020urvos}, CMPC-V \cite{liu2021cross}, LBDT \cite{ding2022language}, MTTR \cite{botach2022end}, ReferFormer \cite{wu2022language}, SgMg \cite{Miao_2023_ICCV}, R2VOS \cite{li2023robust} and OnlineRefer \cite{wu2023onlinerefer}. 
In Table~\ref{tab3}, we report the quantitative performance of our network and eight referring video segmentation methods on the RVSD datasets. 
According to Table~\ref{tab3}, we can find that our RSM-Net clearly outperforms all compared methods, since our RSM-Net achieves the highest scores across nearly all evaluated metrics. 
Moreover, the last row ("Gain") in Table~\ref{tab3} highlights our method's improvements over competing approaches, and our network has an Overall IOU increase of 4.4\%.

Moreover, Fig.~\ref{fig:demo} shows the visual comparisons of RVSD results produced by our network and other state-of-the-art methods.
Apparently, with the same text descriptions and input video frames, our network can better identify the referred shadow regions than the compared methods, and our results are more consistent with the ground truths.
For example, the first, second, and third rows of Fig.~\ref{fig:demo} show that our method can accurately segment the described region when there are multiple shadow regions present, whereas other methods (MTTR \cite{botach2022end} in the second and third rows) may segment irrelevant shaded regions or may fail to recognize the corresponding shaded region (ReferFormer \cite{wu2022language} and SgMg \cite{Miao_2023_ICCV} in the first row). The fourth and fifth rows illustrate the segmentation for the shadow of interest through text descriptions when the object is not entirely visible or absent in the image, a feat not achievable with current instance shadow segmentation techniques. Our results can be observed to more closely align with the ground truths.

\subsection{Ablation Studies}
We further conduct ablation studies to validate the effectiveness of our MSA and TSM designs. To do so, we construct a baseline (denoted as "Baseline") by eliminating our MSA and TSM from our RSM-Net. Subsequently, we incrementally integrate MSA and TSM into the "Baseline" to formulate four networks, which are denoted as "M1", "M2", "M3", and our RSM-Net. 
Table~\ref{tab4} reports the quantitative results of these networks.
Apparently, "M1" outperforms "Baseline", showcasing the effectiveness of the coarse shadow mask attention in our MSA for RVSD.
The improvement in metrics from "M2" to "M1" demonstrates the contribution of the intra-clip memory features in our network.
The progression from "M2" to "M3" with greater metric results further suggests that the incorporation of inter-clip features at a single temporal scale can bolster RVSD.
while the larger metric results of "M3" than "M2", which further indicates that the inter-clip features at a single temporal scale can also enhance RVSD.
In the end, our network has a superior performance over "M3", which demonstrates our hierarchical inter-clip features enable a better RVSD performance in our network.

\section{Conclusion} \label{conclusion}
In this study, we pioneer the RVSD task, which integrates linguistic prompts with video shadow detection, paving the way for new potential applications, such as interactive video editing.
Our first contribution is the development and annotation of the dataset for RVSD, comprising 86 videos paired with 15,011 text descriptions and corresponding shadow masks.
Furthermore, we devise a Twin-Track Synergistic Memory (TSM) module to learn intra-clip and hierarchical inter-clip memory features to boost segmentation performance and a Mixed-Prior Shadow
Attention (MSA) module to learn a coarse shadow attention map for refining shadow areas for RVSD.
Experimental results demonstrate that our method achieves better performance than other leading comparative methods.

%%
%% The acknowledgments section is defined using the "acks" environment
%% (and NOT an unnumbered section). This ensures the proper
%% identification of the section in the article metadata, and the
%% consistent spelling of the heading.
\begin{acks}
This work was supported by the Guangzhou-HKUST(GZ) Joint Funding Program (No. 2023A03J0671), the InnoHK funding launched by Innovation and Technology Commission, Hong Kong SAR, the Guangzhou Industrial Information and Intelligent Key Laboratory Project (No. 2024A03J0628), the Nansha Key Area Science and Technology Project (No. 2023ZD003), and Guangzhou-HKUST(GZ) Joint Funding Program (No. 2024A03J0618).
\end{acks}

%%
%% The next two lines define the bibliography style to be used, and
%% the bibliography file.
\bibliographystyle{ACM-Reference-Format}
\balance
\bibliography{sample-base}

\end{document}